\documentclass[11pt]{article}

\usepackage[utf8]{inputenc}
\usepackage[T1]{fontenc}
\usepackage{lmodern}  
\usepackage{amsmath,amssymb,amsthm}
\usepackage{graphicx}
\usepackage{booktabs}
\usepackage{multirow}
\usepackage{pifont}
\usepackage[hypertexnames=false]{hyperref}
\usepackage{url}
\usepackage{xspace}
\usepackage{microtype}
\usepackage{listings}
\usepackage{xcolor}
\usepackage{tcolorbox}
\usepackage{tikz}
\usetikzlibrary{shapes.geometric, arrows.meta, positioning}
\usepackage{algorithm}
\usepackage{algpseudocode}
\usepackage[margin=1.0in]{geometry}
\hypersetup{
    colorlinks=true,
    linkcolor=blue,
    filecolor=magenta,
    urlcolor=blue,
    citecolor=blue,
    pdftitle={OpenGloss: A Synthetic Encyclopedic Dictionary and Semantic Knowledge Graph},
    pdfauthor={Michael J. Bommarito II},
    pdfsubject={Knowledge Graphs, Natural Language Processing, Semantic Networks},
    pdfkeywords={knowledge graph, wordnet, semantic network, encyclopedic dictionary, synthetic data},
    breaklinks=true,
    plainpages=false,
    pdfstartview=FitH
}

\newcommand{\opengloss}{OpenGloss\xspace}

\newcommand{\cmark}{\ding{51}}
\newcommand{\xmark}{\ding{55}}

\title{OpenGloss: \\ A Synthetic Encyclopedic Dictionary \\and Semantic Knowledge Graph}

\author{
Michael J. Bommarito II\thanks{Portions of this work were prepared with assistance from large language models. The author is solely responsible for all content, including any errors or omissions.} \\
\texttt{michael.bommarito@gmail.com}
}

\date{\today}

\begin{document}

\maketitle

\begin{abstract}
We present \textbf{OpenGloss}, a synthetic encyclopedic dictionary and semantic knowledge graph for English that integrates lexicographic definitions, encyclopedic context, etymological histories, and semantic relationships in a unified resource. OpenGloss contains \textbf{537K senses across 150K lexemes}, on par with WordNet 3.1 and Open English WordNet, while providing \textbf{more than four times as many sense definitions}. These lexemes include \textbf{9.1M semantic edges, 1M usage examples, 3M collocations, and 60M words of encyclopedic content}.

Generated through a \textbf{multi-agent procedural generation pipeline} with schema-validated LLM outputs and automated quality assurance, the entire resource was produced in under one week for under \$1,000. This demonstrates that structured generation can create comprehensive lexical resources at cost and time scales impractical for manual curation, enabling rapid iteration as foundation models improve. The resource addresses gaps in pedagogical applications by providing integrated content---definitions, examples, collocations, encyclopedias, etymology---that supports both vocabulary learning and natural language processing tasks.

As a synthetically generated resource, OpenGloss reflects both the capabilities and limitations of current foundation models. The dataset is publicly available \href{https://huggingface.co/datasets/mjbommar/opengloss-dictionary-definitions}{on Hugging Face} under CC-BY 4.0, enabling researchers and educators to build upon and adapt this resource.
\end{abstract}

\section{Introduction}
\label{sec:introduction}

Every dictionary entry is an opportunity for a human or machine to find the meanings, histories, and relationships needed to understand a word in context. Ideally, lexical resources bridge human comprehension and computational reasoning by providing definitions, usage examples, encyclopedic notes, etymology, and semantic relationships within a single schema. Yet that ideal has been continually deferred; existing resources face unavoidable trade-offs between quality, coverage, currency, and cost.

WordNet\footnote{WordNet is a registered trademark of Princeton University.}~\cite{miller1995wordnet,fellbaum1998wordnet} exemplifies manual curation. It contains 117{,}000 synsets with expert-validated semantic relations, but its update cycles have stagnated since Princeton's final 2011 release. BabelNet~\cite{navigli2012babelnet} pursues multilingual breadth (23 million synsets across 600 languages) by stitching together many sources, yet inherits their uneven coverage and schema misalignments. ConceptNet~\cite{speer2017conceptnet} lowers the cost of creation through crowdsourcing, but schema consistency and quality control remain persistent hurdles. These trade-offs ripple into downstream applications like sense disambiguation, controllable generation, and educational technology, where scale, consistency, and pedagogical richness are required.

Recent breakthroughs in foundation models and structured generation make a new approach plausible. Large language models capture and encode knowledge spanning definitions, semantic relationships, and etymological narratives. Structured generation tools such as \texttt{pydantic-ai}~\cite{pydanticai2024} enable extraction of that knowledge into consistently typed, validated schemas at scale. Prior work has shown that LLMs can produce domain knowledge graphs~\cite{bosselut2019comet,chen2023autokg,hubert2024pygraft}, bootstrap resources for under-resourced languages~\cite{yong2024lexcgen,velasco2023towards}, and perform zero-shot semantic tasks~\cite{arun2025semma,bai2025autoschemakg}. We present \opengloss, a synthetic encyclopedic dictionary and semantic knowledge graph demonstrating that such approaches can now be scaled to create systematic, validated resources at the scale of traditional lexical resources---at practical cost and time scales.

\subsection*{Contributions and Design}

This work makes three primary contributions:

\textbf{First}, we provide a large-scale synthetic lexical dataset with 537{,}000 sense definitions across 150{,}000 lexemes---comparable to WordNet 3.1 in vocabulary breadth while providing 4.6$\times$ more sense definitions. Each entry integrates multiple content dimensions: encyclopedic context (200--400 words, 99.7\% coverage), etymological histories (97.5\% coverage), usage examples (averaging 2 per sense), collocations (3--6 per part-of-speech), and semantic relationships (9.1 million edges). This pedagogical focus addresses gaps in existing computational resources, which typically provide definitions and semantic relations but lack contextual content that supports vocabulary learning and reading comprehension.

\textbf{Second}, we establish a reproducible methodology for synthetic lexical resource creation through a multi-agent generation pipeline with schema-validated outputs and automated quality assurance (Section~\ref{sec:methodology}). Operating within modest budgets (under \$1{,}000, 96 hours), this approach enables rapid iteration as foundation models improve and makes comprehensive lexical resources accessible to individual research groups without requiring institutional infrastructure.

\textbf{Third}, we provide empirical analysis situating \opengloss within the landscape of lexical resource development (Section~\ref{sec:comparative_analysis}). Comparisons with WordNet, BabelNet, and ConceptNet reveal complementary rather than redundant coverage: \opengloss and WordNet share only 38\% vocabulary overlap, with each contributing distinct lexicographic priorities. This analysis examines trade-offs across development approaches---manual curation, integration, crowdsourcing, and systematic generation.

Following WordNet's pragmatic philosophy of prioritizing practical utility~\cite{miller1995wordnet,fellbaum1998wordnet}, \opengloss adopts design choices favoring usability: inclusion of inflected forms and proper nouns enables lookup as encountered; sense granularity (averaging 3.6 senses per lexeme, constrained to 1--4 per part-of-speech) balances comprehensibility with computational tractability. These choices reflect a focus on educational technology and general NLP applications rather than tasks requiring fine-grained semantic precision.

The dataset is publicly available on Hugging Face under CC-BY 4.0 at both the lexeme level\footnote{\url{https://huggingface.co/datasets/mjbommar/opengloss-dictionary}} and sense level\footnote{\url{https://huggingface.co/datasets/mjbommar/opengloss-dictionary-definitions}}. As a synthetically generated resource, \opengloss reflects both capabilities and limitations of current foundation models; Section~\ref{sec:discussion} discusses quality profiles, validation results, and appropriate use cases in detail.

\section{Related Work}
\label{sec:related_work}

Our work builds upon several research areas including traditional lexical databases, semantic networks, knowledge base construction, and recent advances in LLM-based knowledge generation. We review each area and position \opengloss within this landscape.

\subsection{Traditional Lexical Databases}

WordNet~\cite{miller1995wordnet,fellbaum1998wordnet} established the foundational English lexical database, organizing approximately 117,659 synsets (synonym sets) into a hierarchical taxonomy connected by semantic relations including hypernymy, meronymy, and antonymy. Created through years of expert lexicographer effort at Princeton University, WordNet became the gold standard for computational semantics and enabled countless applications in word sense disambiguation, semantic similarity, information extraction, and knowledge-enhanced systems. WordNet's manual curation ensures exceptional quality and taxonomic precision, with expert effort concentrated on core vocabulary with carefully vetted semantic relations. This focus on quality established the standard for structured lexical resources and demonstrated the value of precise sense distinctions for computational applications.

Specialized resources complement WordNet's general-purpose coverage by demonstrating the depth achievable through focused curation. FrameNet~\cite{baker1998berkeley,ruppenhofer2016framenet} documents semantic frames representing prototypical situations with associated participants and roles, providing rich event semantics for approximately 1,200 frames covering 13,000 lexical units. VerbNet~\cite{kipper2008large} focuses on verb semantics, organizing 6,500 verbs into 270 classes based on syntactic behavior and semantic properties. PropBank~\cite{palmer2005proposition} annotates predicate-argument structures in corpus data, creating a valuable resource for semantic role labeling. These resources offer extraordinary semantic depth within their domains, establishing models for how detailed linguistic analysis can inform computational systems.

Multilingual resources extend the WordNet model across languages through alignment. EuroWordNet~\cite{vossen1998eurowordnet} pioneered cross-lingual semantic networks, while the Open Multilingual WordNet~\cite{bond2013linking,bond2016open} provides aligned wordnets for 150+ languages through the Collaborative Interlingual Index. These efforts demonstrate the universal value of structured lexical resources while creating opportunities for approaches that can complement expert curation with broader coverage.

\subsection{Integration-Based and Crowdsourced Resources}

Building on traditional lexical databases, researchers developed complementary approaches that aggregate existing resources or crowdsource knowledge from distributed contributors. BabelNet~\cite{navigli2012babelnet} exemplifies the integration strategy, combining WordNet, Wikipedia, Wiktionary, Wikidata, and other dictionaries through automatic extraction and alignment. BabelNet version 5.3 contains approximately 23 million synsets spanning 600 languages, achieving unprecedented multilingual coverage. This massive scale demonstrates the power of strategic resource combination: Wikipedia articles provide encyclopedic content, Wiktionary supplies multilingual translations, and WordNet anchors the semantic network with curated precision.

Integration approaches balance scale and precision through strategic design choices. Automatic alignment methods enable rapid coverage expansion, with performance particularly strong for concepts with substantial Wikipedia coverage. This demonstrates how strategic integration can leverage existing human effort. BabelNet's approach naturally inherits the coverage patterns of its constituent resources, combining their complementary strengths.

Crowdsourcing offers a complementary strategy for knowledge acquisition. ConceptNet~\cite{speer2012conceptnet,speer2017conceptnet} aggregates commonsense knowledge from multiple sources including crowd-contributed databases (Open Mind Common Sense), games with a purpose (Verbosity), and automatic extraction from web text. Version 5.7 contains approximately 21 million edges connecting 8 million concepts across 83 languages. ConceptNet's relationship types (``UsedFor,'' ``Causes,'' ``LocatedAt,'' ``CapableOf'') capture everyday reasoning patterns complementary to lexicographic resources: while WordNet excels at taxonomic precision (``a dog is a type of canine''), ConceptNet captures functional knowledge (``dogs are used for companionship,'' ``dogs are capable of barking''). This commonsense focus makes ConceptNet particularly valuable for reasoning applications, demonstrating how different knowledge representations serve different computational needs.

\subsection{Knowledge Base Construction and Extraction}

Automatic knowledge base construction methods extract structured knowledge from unstructured text at scale. Early systems like NELL~\cite{carlson2010toward} employed bootstrapping to continuously read web pages and extract factual assertions, demonstrating the potential for never-ending learning. Relation extraction methods identify entity relationships in text using pattern matching, supervised learning, or distant supervision. Open Information Extraction (Open IE)~\cite{banko2007open} extracts relation triples without predefined schemas, enabling broad coverage through domain-independent methods.

Recent neural approaches apply transformer models to knowledge base completion, link prediction, and entity alignment. Knowledge graph embedding methods like TransE~\cite{bordes2013translating}, DistMult~\cite{yang2015embedding}, and ComplEx~\cite{trouillon2016complex} learn vector representations of entities and relations that enable reasoning over incomplete graphs. These techniques excel at scaling and enriching existing knowledge bases, demonstrating how neural methods can complement symbolic representations.

\subsection{LLM-Based Knowledge Generation}

Large language models represent a natural evolution in knowledge generation capabilities. KG-BERT~\cite{yao2019kgbert} and REBEL~\cite{cabot2021rebel} pioneered treating knowledge graph tasks as sequence modeling. BertNet~\cite{hao2023bertnet} extracts symbolic knowledge graphs from pretrained BERT, while TaxoLLaMA~\cite{moskvoretskii2024taxollama} fine-tunes LLaMA-2 on WordNet data for lexical semantic tasks. Foundation models like Semma~\cite{arun2025semma} and AutoSchemaKG~\cite{bai2025autoschemakg} enable zero-shot link prediction and automatic ontology induction, demonstrating how pretrained models encode rich linguistic and world knowledge.

Frameworks for structured generation have emerged as critical infrastructure for reliable knowledge production. \texttt{pydantic-ai}~\cite{pydanticai2024} provides type-safe agents with schema validation, while constrained decoding methods~\cite{hokamp2017lexically,willard2023efficient} ensure outputs conform to formal grammars. GraphRAG~\cite{edge2024graphrag}, Text2KGBench~\cite{mihindukulasooriya2023text2kgebench}, and iText2KG~\cite{lairgi2024itext2kg} demonstrate hierarchical and incremental knowledge graph construction from documents.

Commonsense generation systems like COMET~\cite{bosselut2019comet} and ATOMIC-10x~\cite{west2022symbolic}, along with domain-specific approaches including PyGraft~\cite{hubert2024pygraft} and AutoKG~\cite{chen2023autokg}, have demonstrated LLM capabilities for generating structured knowledge in focused domains.

\subsection{Multilingual and Hybrid Approaches}

Multilingual resources like mBERT~\cite{devlin2019bert} and XLM-RoBERTa~\cite{conneau2020unsupervised} enable cross-lingual transfer, extending language understanding across linguistic boundaries. Recent work including LexC-Gen~\cite{yong2024lexcgen} and FilWordNet~\cite{borra2010filwordnet,velasco2023towards} demonstrates that LLMs can bootstrap lexical databases for low-resource languages. Hybrid neural-symbolic architectures like KEPLER~\cite{wang2021kepler} and CoLAKE~\cite{sun2020colake} combine language modeling with knowledge graph embeddings. Educational applications~\cite{tarus2018knowledge} leverage structured semantic knowledge for personalized learning and curriculum design.

\subsection{Positioning OpenGloss}

\opengloss addresses gaps left by existing approaches through systematic LLM-based generation of comprehensive lexical resources. Where WordNet achieves exceptional quality through manual curation but at limited scale and slow update frequency, \opengloss now surpasses WordNet's lexeme coverage (1.02$\times$) with 4.59$\times$ more sense definitions, demonstrating systematic validation at accessible cost and time scales. BabelNet's integration strategy achieves massive multilingual coverage (23M synsets across 600 languages) but inherits heterogeneity across constituent sources. In contrast, \opengloss generates original content with consistent structure through schema-validated generation. Unlike ConceptNet, which captures commonsense relationships (21M edges, 8M concepts) optimized for functional knowledge over lexicographic completeness, \opengloss provides comprehensive lexical coverage combining definitions, encyclopedic context, etymological histories, and semantic relationships in a unified resource. Focused LLM approaches like COMET, AutoKG, and PyGraft have demonstrated domain-specific knowledge generation capabilities. \opengloss extends these insights to general lexical resource creation through multi-agent generation with automated quality assurance.

This positioning reflects three novel contributions relative to existing work. First, \textbf{comprehensive scope}: \opengloss is the first LLM-generated resource to combine lexicographic definitions, encyclopedic context, etymological histories, and semantic networks at scale, demonstrating that systematic generation can produce integrated lexical resources rather than specialized fragments. Second, \textbf{systematic validation}: the multi-agent pipeline with schema-validated outputs and automated quality assurance provides reproducible methodology for lexical resource creation, enabling rapid iteration as foundation models improve. Third, \textbf{practical accessibility}: achieving comprehensive coverage at cost (\textless\$1,000) and time scales (96 hours) feasible for individual research groups demonstrates that LLM-based generation complements rather than replaces existing approaches---manual curation for gold-standard quality, integration for multilingual coverage, crowdsourcing for commonsense knowledge, and now systematic generation for comprehensive lexical resources with rapid update cycles.

\section{Methodology}
\label{sec:methodology}

This section describes the comprehensive pipeline used to generate \opengloss, from lexeme selection through final validation. Our approach combines multi-agent LLM generation with rigorous schema validation to produce structured lexical data at scale while maintaining consistency and quality.

\subsection{Pipeline Overview}

\begin{figure}[t]
\centering
\begin{tcolorbox}[
    colback=gray!2,
    colframe=gray!50,
    arc=2mm,
    boxrule=0.5pt,
    left=4pt,
    right=4pt,
    top=4pt,
    bottom=4pt,
    width=0.98\columnwidth,
    halign=center
]
\centering
\begin{tikzpicture}[
    node distance=3.5cm,
    stage/.style={rectangle, draw=black, very thick,
                  text width=2cm, minimum height=1.4cm, align=flush center, font=\small},
    data/.style={rectangle, draw=black, fill=black!5,
                 text width=1.8cm, minimum height=0.85cm, align=flush center, font=\small},
    arrow/.style={->, >=stealth, very thick},
    note/.style={font=\scriptsize, align=center}
]

\node[stage] (s1) {\textbf{Stage 1}\\[0.2em]Lexeme\\Selection};
\node[stage, right of=s1] (s2) {\textbf{Stage 2}\\[0.2em]Sense\\Generation};
\node[stage, right of=s2] (s3) {\textbf{Stage 3}\\[0.2em]Graph\\Construction};
\node[stage, right of=s3] (s4) {\textbf{Stage 4}\\[0.2em]Enrichment};

\node[data, below=1.8cm of s1] (d1) {150K\\lexemes};
\node[data, below=1.8cm of s2] (d2) {537K\\senses};
\node[data, below=1.8cm of s3] (d3) {9.1M\\edges};
\node[data, below=1.8cm of s4] (d4) {Final\\Dataset};

\draw[arrow] (s1) -- (s2);
\draw[arrow] (s2) -- (s3);
\draw[arrow] (s3) -- (s4);

\draw[arrow] (s1) -- (d1);
\draw[arrow] (s2) -- (d2);
\draw[arrow] (s3) -- (d3);
\draw[arrow] (s4) -- (d4);

\draw[arrow, dashed, rounded corners=5pt]
    (d3.south) -- ++(0,-0.5) -| (d1.south);

\node[note] at (4.0,-4.5) {snowball sampling};

\node[data, below=1.5cm of d4] (qa) {QA\\1K sample};
\draw[arrow, dashed] (d4) -- (qa);

\end{tikzpicture}
\end{tcolorbox}
\caption{OpenGloss generation pipeline. The four-stage process combines multi-agent LLM generation with deterministic graph construction and systematic enrichment. Pydantic schema validation ensures type safety at each stage. The entire pipeline completed in 96 hours at under \$1,000 using gpt-5-nano, with automated QA using Claude Sonnet 4.5.}
\label{fig:methodology_pipeline}
\end{figure}
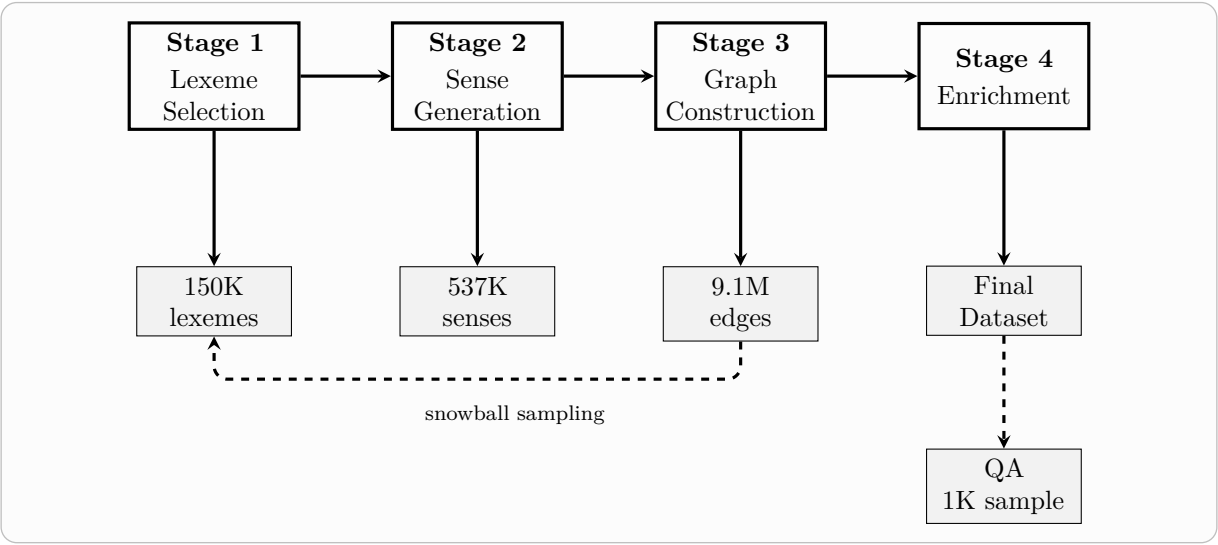

Figure~\ref{fig:methodology_pipeline} illustrates our four-stage generation pipeline. \opengloss{} was constructed using \texttt{pydantic-ai}~\cite{pydanticai2024} for type-safe LLM interactions with Pydantic V2 schema validation. Each stage---from lexeme selection through final enrichment---validates outputs against Pydantic schemas, enabling modular fault isolation and systematic quality control. The pipeline incorporates a snowball sampling feedback loop where graph construction reveals new pedagogically relevant lexemes for inclusion. Detailed statistics and stage descriptions follow in subsequent subsections.

\subsection{Lexeme Selection}
\label{sec:methodology:lexeme-selection}

\textbf{OpenGloss's original focus is pedagogical: supporting K-12 education} by providing comprehensive lexical resources for vocabulary learning, reading comprehension, and educational content development. Rather than attempting comprehensive English coverage---an impossibly large task given productive morphology and neologisms---we prioritized vocabulary students encounter in educational contexts and everyday communication. We selected 150,101 lexemes from two primary seed sources.

First, we began with the \texttt{wamerican} package dictionary\footnote{Specifically, version 2020.12.07-4 at path \texttt{/usr/share/dict/words}.}, containing 104,334 words representing standard American English vocabulary. From this source, we applied minimal filtering: (1) length constraints (3-15 characters) to exclude abbreviations and extremely long technical terms, (2) alphabetic-only filtering to remove possessive forms (e.g., ``Aaron's''), and (3) deduplication. Notably, we \emph{included} proper nouns (e.g., ``Aaron,'' ``London,'' ``Einstein'') and inflectional variants when pedagogically relevant, resulting in 73,200 words from \texttt{wamerican} (70.2\% coverage). Second, we augmented this foundation by traversing an LLM-proposed neighbor graph seeded with everyday objects, concepts, and K-12 educational topics. Starting from seed concepts, we iteratively expanded to related terms, synonyms, hypernyms, and domain-specific vocabulary, adding 76,901 lexemes not found in \texttt{wamerican}. This iterative expansion---the snowball sampling feedback loop shown in Figure~\ref{fig:methodology_pipeline}---allows discovered semantic relationships during graph construction to suggest additional pedagogically relevant lexemes for inclusion.

The two-source strategy combines established lexicographic coverage (73,200 from \texttt{wamerican}) with targeted educational expansion (76,901 additional terms), yielding 150,101 total lexemes. The selection includes both single-word entries (94,106 lexemes, 62.7\%) and multi-word expressions (55,995 lexemes, 37.3\%) such as ``a bit,'' ``agricultural system,'' and ``a lot of homework''---reflecting the multi-word phrases students encounter in textbooks and classroom discourse. The pedagogical focus shapes vocabulary priorities: comprehensive coverage of foundational concepts, everyday objects, academic terminology across subject areas (mathematics, science, social studies, language arts), and developmentally-sequenced vocabulary.

\subsection{Data Model}
\label{sec:methodology:data-model}

\begin{figure}[t]
\centering
\begin{tcolorbox}[
    colback=gray!2,
    colframe=gray!50,
    arc=2mm,
    boxrule=0.5pt,
    left=4pt,
    right=4pt,
    top=4pt,
    bottom=4pt,
    width=0.98\columnwidth,
    halign=center
]
\centering
\begin{tikzpicture}[
    parent/.style={rectangle, draw=black, very thick,
                   text width=2.2cm, minimum height=0.7cm, align=center, font=\small},
    child/.style={rectangle, draw=black, fill=black!5,
                  text width=2.2cm, minimum height=0.7cm, align=center, font=\small},
    arrow/.style={->, >=stealth, thick}
]

\node[parent] (lexeme) at (0,0) {\textbf{Lexeme}};
\node[parent] (pos) at (0,-2.5) {\textbf{Part of Speech}};
\node[parent] (sense) at (0,-5) {\textbf{Lexical Sense}};

\draw[arrow] (lexeme) -- (pos);
\draw[arrow] (pos) -- (sense);

\node[child] (encyc) at (3.5,0) {Encyclopedia};
\node[child, below=0cm of encyc] (etym) {Etymology};
\draw[arrow] (lexeme) -- (encyc);

\node[child] (morph) at (3.5,-2.5) {Morphology};
\node[child, below=0cm of morph] (colloc) {Collocations};
\draw[arrow] (pos) -- (morph);

\node[child] (syn) at (3.5,-5) {Synonyms};
\node[child, below=0cm of syn] (ant) {Antonyms};
\node[child, below=0cm of ant] (hyper) {Hypernyms};
\node[child, below=0cm of hyper] (hypo) {Hyponyms};
\node[child, below=0cm of hypo] (ex) {Examples};
\draw[arrow] (sense) -- (syn);

\end{tikzpicture}
\end{tcolorbox}
\caption{OpenGloss data model hierarchy. The Pydantic schema organizes information at three levels: Lexeme (root container with etymology and encyclopedia), Part of Speech (POS-specific with 1-4 senses and morphology), and Lexical Sense (atomic unit with definition and semantic neighborhood). The hierarchical structure supports both computational access and traditional lexicographic organization.}
\label{fig:data_model}
\end{figure}
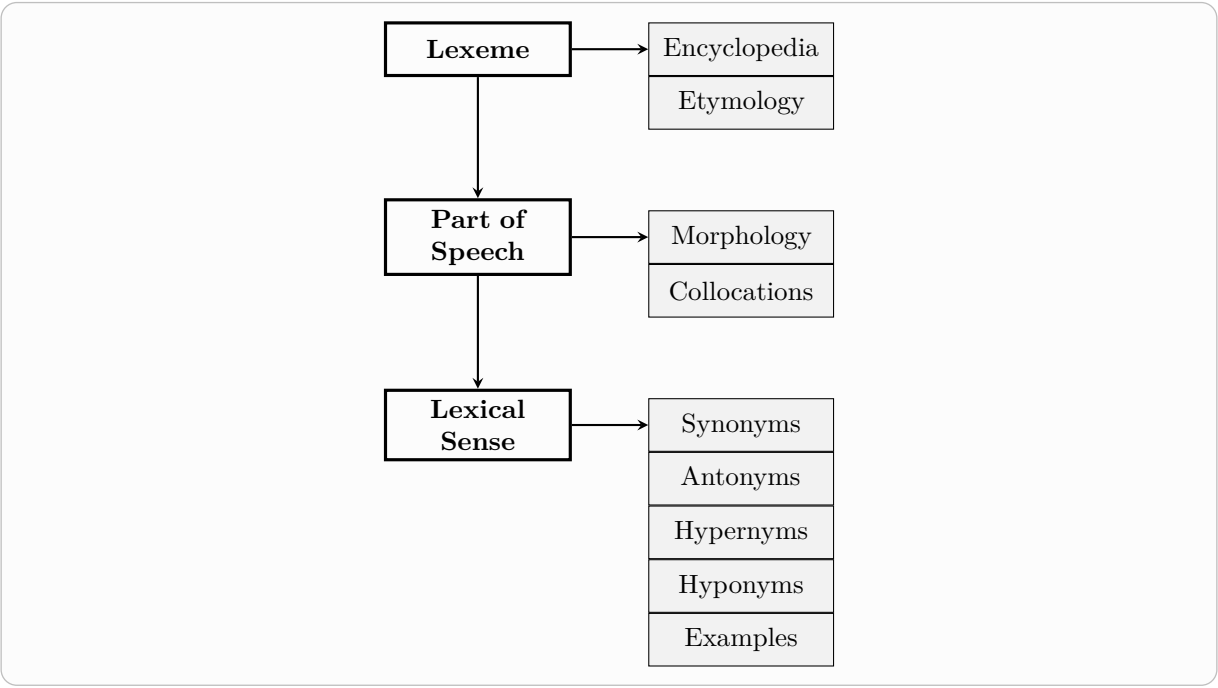

The foundation of structured LLM generation is a well-designed data model that constrains outputs while remaining flexible enough to capture linguistic complexity. Figure~\ref{fig:data_model} illustrates our hierarchical Pydantic schema with a vertical spine representing containment (Lexeme contains Part-of-Speech entries, which contain Lexical Senses) and horizontal branches showing information attached at each level: lexeme-level attributes (encyclopedia, etymology), POS-level attributes (morphology, collocations), and sense-level semantic relationships (synonyms, antonyms, hypernyms, hyponyms, examples). The following subsections detail the core components and their hierarchical relationships, emphasizing aspects not immediately visible in the diagram.

\subsubsection{LexemeEntry: Root Model}

The \texttt{LexemeEntry} serves as the root container (top of the vertical spine in Figure~\ref{fig:data_model}) for all information about a single lexeme (which may be a single word or multi-word expression). As shown in the figure, each lexeme entry contains etymology and encyclopedic information as horizontal leaf nodes, while also serving as the parent container for one or more part-of-speech entries through the vertical hierarchy. Beyond these hierarchical relationships visualized in the figure, the root model also maintains lexeme-level semantic edges (e.g., morphological derivations like ``happy'' $\rightarrow$ ``happiness''), stopword classification with reasoning, and generation metadata (timestamp, schema version, validation flags).

This root-level organization mirrors traditional dictionary structure while supporting computational access: applications can query by lemma, filter by part-of-speech, traverse semantic relationships, or access encyclopedic context as needed.

\subsubsection{PartOfSpeechEntry: POS-Specific Data}

Each \texttt{PartOfSpeechEntry} (middle level in Figure~\ref{fig:data_model}) captures information specific to one grammatical category. The POS label uses an enumerated type from a fixed taxonomy (noun, verb, adjective, adverb, determiner, preposition, conjunction, pronoun, interjection). As the figure shows, each POS entry contains morphology and collocation information as horizontal leaf attributes, while maintaining 1-4 \texttt{LexicalSense} objects representing distinct meanings within this grammatical category through the vertical hierarchy. The 1-4 constraint balances sense granularity with generation tractability: too few senses miss important distinctions, while too many introduce spurious over-specification.

Organizing data by part-of-speech facilitates precise querying and avoids sense conflation: the noun and verb senses of ``bank'' are semantically unrelated and benefit from separate treatment. This mirrors traditional lexicographic practice while enabling computational applications that require POS disambiguation.

\subsubsection{LexicalSense: Definition and Semantic Neighborhood}

The \texttt{LexicalSense} (bottom of the hierarchy in Figure~\ref{fig:data_model}) represents a single word meaning---the atomic unit of lexical semantics. Each sense consists of a concise prose definition (typically 50-150 characters) following lexicographic conventions, plus the semantic neighborhood shown as horizontal leaf nodes in the figure: synonyms (3-5 closely related words), antonyms (contrasting meanings when applicable), hypernyms (broader concepts like ``canine'' for ``dog''), hyponyms (narrower concepts like ``apple'' for ``fruit''), and 1-3 usage examples demonstrating typical grammatical context.

Such semantic neighborhoods support graph-based reasoning: applications can traverse hypernym chains to find conceptual generalizations, identify semantically similar concepts through shared hypernyms, or generate explanations by following hyponym edges to concrete examples.

\subsubsection{Morphology and Semantic Edges}

The \texttt{Morphology} component captures systematic form variation through both \textbf{inflections} (grammatically required variants such as noun plurals, verb tenses, and adjective comparison) and \textbf{derivations} (cross-category morphological relationships like ``happy'' $\rightarrow$ ``happiness'' or ``nationalize'' $\leftarrow$ ``national'' $\leftarrow$ ``nation''). Morphological information supports linguistic analysis, allows generation systems to produce grammatically correct forms, and helps NLP applications handle unseen inflections.

The \texttt{LexemeEdge} type encodes directed, typed relationships between lexemes or senses. We define 13 relationship types spanning three categories: \textbf{sense-level} relations (synonym, antonym, hypernym, hyponym) capture paradigmatic semantic patterns; \textbf{lexeme-level} relations include syntagmatic collocations and morphological derivations (inflection, noun-derivation, verb-derivation, adjective-derivation, adverb-derivation); and \textbf{historical} relations document cross-linguistic cognates, morpheme components, and etymology-parent precursor forms. Each edge includes source and target identifiers, relationship type, and optional relationship strength or frequency. This explicit typing enables precise graph queries and supports diverse applications requiring different relationship types.

\subsection{Multi-Agent Generation Pipeline}
\label{sec:methodology:pipeline}

The generation pipeline employs four specialized LLM agents, each designed for a specific subtask. The multi-agent architecture enables task-specific prompting, reduces cognitive load per agent, and facilitates quality control by isolating failure modes. All agents use \texttt{pydantic-ai}'s type-safe \texttt{Agent} class with configurable LLM models as the underlying model.

For the \opengloss v1.0 release described in this paper, we instantiated this pipeline with OpenAI's \texttt{gpt-5-nano} model via the \texttt{pydantic-ai} interface. Because the pipeline is backend-agnostic, practitioners can substitute different models or deployment environments while preserving the same structured generation process.

\subsubsection{Stage 1: Lexeme Selection}

The first stage (Figure~\ref{fig:methodology_pipeline}) establishes the vocabulary foundation. As described in Section~\ref{sec:methodology:lexeme-selection}, we selected 150,101 lexemes from frequency-based American English word lists (\texttt{wamerican}), prioritizing K-12 and college educational vocabulary. This vocabulary list shapes all subsequent generation stages, as
snowball sampling builds off the initial network structure.

\subsubsection{Stage 2: Sense Generation}

The sense generation stage (Figure~\ref{fig:methodology_pipeline}) employs a two-agent architecture that balances broad coverage with detailed semantic information. The overview agent determines valid part-of-speech categories, stopword classification, and approximate sense counts per POS. For each valid POS category, the POS details agent generates 1-4 sense definitions with semantic relationships (3-5 synonyms, antonyms when applicable, 2-4 hypernyms/hyponyms, 1-3 usage examples), morphology (inflectional and derivational forms), and 3-6 collocations. The agent returns structured Pydantic models validated immediately---malformed outputs are rejected. The two-stage process produced 536,829 sense definitions across 150,101 lexemes (3.58 senses/word average), matching WordNet's polysemy patterns while extending synset coverage 4.59-fold.

\subsubsection{Stage 3: Graph Construction}

As shown in Figure~\ref{fig:methodology_pipeline}, graph construction transforms the flat semantic relationships embedded in individual senses into an explicit, queryable semantic network. This stage operates deterministically (without LLM calls) by extracting edges from structured sense data:

\paragraph{Sense-Level Edges} For each sense, we extract four relationship types: \textbf{synonym edges} connect senses with highly similar meanings to create synonym clusters; \textbf{antonym edges} link semantically opposed concepts; \textbf{hypernym edges} create taxonomic hierarchies from specific to general concepts; and \textbf{hyponym edges} (inverse of hypernyms) link general concepts to specific instances. These 5.20 million sense-level edges form the core semantic network supporting applications requiring semantic similarity, taxonomy navigation, and conceptual reasoning.

\paragraph{POS-Level Edges} At the lexeme level, we extract \textbf{collocation edges} connecting lexemes frequently appearing together (3.06 million edges) and \textbf{inflection edges} linking lemmas to inflectional variants (875,673 edges).

\paragraph{Morphological and Etymology Edges} Additional edge types capture linguistic structure: \textbf{derivation edges} connect morphologically related lexemes across POS categories (``happy'' $\leftrightarrow$ ``happiness''), while \textbf{etymology edges} link lexemes to historical precursors (cross-linguistic cognates are stored within etymology metadata rather than as separate graph edges). Edge priority classification helps downstream applications weight relationships appropriately: high priority (hypernyms, hyponyms, antonyms---semantic backbone), medium priority (synonyms, derivations---lexical structure), and low priority (collocations, inflections---surface co-occurrence).

The resulting graph contains 9.14 million edges total, providing rich connectivity for traversal and reasoning.  Notably, as discussed below, our generation process results in a much less right-tailed degree distribution as compared to WordNet.  As a result, while the graph is still large, most graph algorithms have significantly faster runtime on OpenGloss than on WordNet.

\subsubsection{Stage 4: Enrichment}

The final pipeline stage (Figure~\ref{fig:methodology_pipeline}) enriches entries through two additional agents. The etymology agent generates historical development trails for non-stopword lexemes documenting language progression, cognates, semantic evolution, and scholarly citations, achieving 97.5\% coverage. The encyclopedia agent generates 200-400 word entries providing conceptual context, key characteristics, applications, historical development, and related concepts in educational prose, achieving 99.7\% sense coverage. This distinguishes \opengloss from traditional dictionaries offering only concise definitions and from encyclopedias lacking systematic sense-level organization.

\subsection{Implementation}
\label{sec:methodology:implementation}

\subsubsection{Infrastructure and Models}

\textbf{LLM backends}: The generation and QA pipelines use \texttt{pydantic-ai}'s LLM-agnostic \texttt{Agent} interface. Separate agents are configured for overview/POS generation, enrichment (etymology and encyclopedias), and quality assurance. Specific model identifiers are configurable; in the v1.0 configuration we used OpenAI's \texttt{gpt-5-nano} for generation and Anthropic's Claude Sonnet 4.5 for QA, but the methodology is not tied to a particular provider.

\textbf{Sampling parameters} (for the configuration used in this release): temperature 0.7 (overview, POS, etymology) and 0.9 (encyclopedia for stylistic diversity), top-p 0.95, max tokens 2048 (overview/POS) and 4096 (encyclopedia), and frequency penalty 0.3 (to reduce repetition in longer encyclopedia texts).  These parameters were also selected to help reduce the probability of reproducing verbatim sequences from training data.

\textbf{Compute and cost} (v1.0 reference run): Generation used \texttt{gpt-5-nano} via OpenAI API without any local GPU requirements, completing in under 96 hours of wall-clock time with total API spend under \$1{,}000 for the full dataset. The marginal cost to reproduce this data using open weight models is likely substantially lower for those moderate hardware, but we selected these models for general demonstration and accessibility.

\subsubsection{Validation and Error Handling}

\textbf{Schema validation}: Pydantic V2 type checking with strict mode enabled. Malformed outputs (estimated 2-4\% of responses) trigger automatic retry with enhanced prompts specifying exact field requirements.

\textbf{Semantic validation}: 100\% edge target validity enforced---all semantic relationship targets (hypernyms, hyponyms, synonyms, antonyms) must exist in lexeme vocabulary. Invalid edges referencing non-existent lexemes are dropped during post-processing.

\textbf{Graph connectivity}: Automated checks ensure acyclic hypernym/hyponym relationships and symmetric synonym/antonym pairs.

\subsection{Quality Assurance Pipeline}
\label{sec:methodology:qa}

Following generation, we applied automated quality assurance designed to evaluate \opengloss against our core design philosophy: emulating WordNet's pragmatic approach that prioritizes practical utility and computational accessibility over strict lexicographic purity. We developed an automated LLM-based QA pipeline using Claude Sonnet 4.5 that evaluates entries from two perspectives: adherence to traditional lexicographic conventions (to identify genuine quality issues) and successful implementation of WordNet-aligned pragmatic choices (to confirm deliberate design decisions).

\textbf{QA Architecture:} The QA agent evaluates entries across two dimensions. First, core content quality assesses entry structure (headword validity, POS appropriateness), definitional quality (clarity, precision, completeness), encyclopedia quality (factual accuracy, pedagogical utility), and etymology quality (plausibility and educational value). Second, semantic relationship quality validates hypernym/hyponym taxonomic hierarchies, synonym meaning equivalence, and antonym oppositional accuracy. For each entry, the agent assigns a verdict (pass/needs\_review/flagged) with severity-classified issues, creating an audit trail for analysis.

\textbf{Validation Study:} We evaluated 1,000 randomly sampled entries from the full dataset (Table~\ref{tab:qa_results}, Appendix~\ref{appendix:qa}). The results confirm successful implementation of our WordNet-aligned design while identifying targeted opportunities for refinement.

A key finding validates our design philosophy: 38.6\% of entries were flagged for including inflected forms (e.g., ``running,'' ``dogs,'' ``better'') or proper nouns (e.g., ``London,'' ``Einstein''). These are not quality failures but successful replications of WordNet's established lexicographic practices. Traditional purist dictionaries exclude these entries to minimize redundancy, but WordNet deliberately includes them to support computational applications where users query the forms they encounter in text. Our QA agent, applying strict traditional standards, flagged these as structural deviations---but as detailed in Section~\ref{sec:lexicographic_philosophy}, they represent positive confirmation of our pragmatic design choices.

Core content quality proved robust: definitions, encyclopedic paragraphs, etymologies, and usage examples were generally usable, confirming the dataset's suitability for educational and general NLP applications. The remaining flags primarily highlight opportunities for iterative refinement in semantic relationship precision (e.g., borderline hypernyms or synonyms). As discussed in Section~\ref{sec:lexicographic_philosophy}, manual review suggests many semantic flags reflect conservative QA standards exceeding WordNet's own practices rather than catastrophic errors, though genuine improvements remain possible.

The QA methodology is generalizable: other projects generating synthetic structured data can adapt this dual-perspective framework---evaluating both against traditional standards and against pragmatic design goals---for validation needs.

\section{Dataset Statistics}
\label{sec:dataset_statistics}

This section presents empirical analysis of the OpenGloss dataset, examining scale, lexical coverage, semantic structure, and content characteristics.

\subsection{Overview Statistics}
\label{sec:overview-statistics}

OpenGloss contains 150,101 lexical entries spanning 536,829 distinct senses, with an average of 3.58 senses per lexeme (Table~\ref{tab:dataset_overview}).

\paragraph{Lexeme Composition} As described in Section~\ref{sec:methodology:lexeme-selection}, the 150,101 lexemes comprise 94,106 single-word entries (62.7\%) and 55,995 multi-word expressions (37.3\%). This substantial MWE component reflects everyday phrasal and idiomatic usage patterns.

The semantic network comprises millions of sense-level and POS-level relationships, detailed in the following subsections. Nearly all entries include encyclopedic content (99.7\%) and etymological trails (97.5\%), distinguishing OpenGloss from resources focusing on a single dimension.

Figure~\ref{fig:lexeme_examples} presents two representative lexeme entries from OpenGloss, illustrating the dataset's comprehensive structure. The \textit{algorithm} entry exemplifies technical computational vocabulary with precise sense distinctions and rich semantic relationships, while the \textit{photosynthesis} entry demonstrates scientific educational content including chemical notation and pedagogical framing. Both examples showcase OpenGloss's integration of lexicographic definitions, semantic networks (synonyms, hypernyms, hyponyms), morphological derivations, collocational patterns, usage examples, and encyclopedic context within a unified framework.

\begin{figure}[t]
\centering
\small

\begin{tcolorbox}[
    colback=gray!2,
    colframe=gray!50,
    colbacktitle=gray!15,
    coltitle=black,
    title=\textbf{algorithm} (noun),
    fonttitle=\normalsize\bfseries,
    width=0.96\columnwidth,
    arc=2mm,
    boxrule=0.5pt,
    left=4pt,
    right=4pt,
    top=4pt,
    bottom=4pt
]

\textbf{Sense 1:} A finite, stepwise procedure for solving a problem or completing a computation.

\hspace{1em}\textit{Synonyms:} procedure, method

\hspace{1em}\textit{Hypernyms:} technique, system

\hspace{1em}\textit{Example:} ``The student traced each algorithm step\ldots''

\vspace{0.3em}

\textbf{Sense 2:} A set of precise rules used to generate a predictable output from given inputs.

\hspace{1em}\textit{Synonyms:} rule, formula

\hspace{1em}\textit{Hypernyms:} framework

\hspace{1em}\textit{Example:} ``Learners tested the algorithm on new inputs\ldots''

\vspace{0.3em}

\textbf{Encyclopedia excerpt:} An algorithm is a finite, well-defined sequence of steps designed to solve a problem or accomplish a specific task. This concept is central to computer science, mathematics, and data analysis. Algorithms underpin everything from basic arithmetic procedures to complex decision-making systems\ldots

\end{tcolorbox}

\vspace{0.3em}

\begin{tcolorbox}[
    colback=gray!2,
    colframe=gray!50,
    colbacktitle=gray!15,
    coltitle=black,
    title=\textbf{photosynthesis} (noun),
    fonttitle=\normalsize\bfseries,
    width=0.96\columnwidth,
    arc=2mm,
    boxrule=0.5pt,
    left=4pt,
    right=4pt,
    top=4pt,
    bottom=4pt
]

\textbf{Sense 1:} The process by which green plants convert light energy into chemical energy, releasing oxygen.

\hspace{1em}\textit{Synonyms:} biosynthesis

\hspace{1em}\textit{Hypernyms:} biological process

\hspace{1em}\textit{Hyponyms:} Calvin cycle

\hspace{1em}\textit{Example:} ``Plants convert light energy into chemical energy during photosynthesis.''

\vspace{0.3em}

\textbf{Encyclopedia excerpt:} Photosynthesis is the biochemical process by which photoautotrophs convert light energy into chemical energy stored in organic molecules. In plants, algae, and cyanobacteria, light energy drives electron transport, producing ATP and NADPH for carbon fixation. The overall stoichiometry: 6 CO$_2$ + 6 H$_2$O + light $\rightarrow$ C$_6$H$_{12}$O$_6$ + 6 O$_2$\ldots

\vspace{0.3em}

\textbf{Etymology:} Greek \textit{photo-} (light) + \textit{synthesis} (putting together), coined 19th century.

\end{tcolorbox}

\caption{Representative lexeme entries from OpenGloss showing core structure: multiple senses with definitions, semantic relationships (synonyms, hypernyms, hyponyms), usage examples, encyclopedic context, and etymology. \textit{Algorithm} represents technical vocabulary; \textit{photosynthesis} illustrates scientific terminology.}
\label{fig:lexeme_examples}
\end{figure}

\begin{table}[t]
\centering
\caption{OpenGloss Dataset Overview}
\label{tab:dataset_overview}
\begin{tabular}{lr}
\toprule
\textbf{Metric} & \textbf{Count} \\
\midrule
Total unique words & 150,101 \\
Total word senses & 536,829 \\
Average senses per word & 3.58 \\
Maximum senses & 24 \\
\midrule
Words with etymology & 146,066 (97.3\%) \\
Words with encyclopedia & 149,614 (99.7\%) \\
\bottomrule
\end{tabular}
\end{table}

\subsection{Lexical Coverage}
\label{sec:lexical-coverage}

\paragraph{Part-of-Speech Distribution}
The dataset exhibits a noun-dominant distribution typical of comprehensive lexical resources, with nouns comprising 51.9\% (278,401) of all senses, followed by adjectives at 26.9\% (144,234) and verbs at 16.9\% (90,664). Function words including adverbs (2.3\%), determiners (0.9\%), prepositions (0.6\%), interjections (0.4\%), pronouns (0.2\%), and conjunctions (0.1\%) account for the remaining 4.5\%. This distribution indicates both natural frequency in English and lexicographic priority on semantically rich open-class categories.

Table~\ref{tab:pos_distribution} presents the complete distribution across all nine part-of-speech categories, showing both sense counts and percentages. This pattern highlights the dominance of nouns in lexical resources and the comparative scarcity of closed-class function words. This pattern aligns with WordNet's distribution and demonstrates the semantic richness of open-class categories: nouns encode entities and concepts, verbs describe events and actions, and adjectives capture properties---all requiring fine-grained sense distinctions. In contrast, function words like determiners and conjunctions serve primarily grammatical roles with limited polysemy.

\begin{table}[t]
\centering
\caption{Part-of-Speech Distribution}
\label{tab:pos_distribution}
\begin{tabular}{lrr}
\toprule
\textbf{Part of Speech} & \textbf{Count} & \textbf{Percentage} \\
\midrule
Noun & 278,401 & 51.9\% \\
Adjective & 144,234 & 26.9\% \\
Verb & 90,664 & 16.9\% \\
Adverb & 12,184 & 2.3\% \\
Determiner & 4,755 & 0.9\% \\
Preposition & 3,379 & 0.6\% \\
Interjection & 1,908 & 0.4\% \\
Pronoun & 857 & 0.2\% \\
Conjunction & 447 & 0.1\% \\
\midrule
\textbf{Total} & \textbf{536,829} & \textbf{100.0\%} \\
\bottomrule
\end{tabular}
\end{table}

\paragraph{Polysemy Patterns}
Polysemy---the phenomenon of lexemes having multiple related meanings---is pervasive in natural language and poses challenges for word sense disambiguation and semantic analysis. OpenGloss exhibits polysemy patterns consistent with corpus linguistics findings: a median of 3 senses per lexeme and a maximum of 24 senses. Of the 150,101 total lexemes, 150,081 (99.99\%) have complete sense definitions, while 20 lexemes (0.01\%) represent incomplete generation edge cases. The distribution of the 150,081 lexemes with senses shows typical characteristics of lexical resources:

\begin{itemize}
    \item \textbf{Monosemous lexemes} (1 sense): 14,233 lexemes (9.5\%)---primarily technical terms, proper nouns, and highly specific concepts
    \item \textbf{Low polysemy} (2-4 senses): 92,264 lexemes (61.5\%)---the majority of vocabulary
    \item \textbf{Moderate polysemy} (5-9 senses): 42,714 lexemes (28.5\%)---common lexemes with extended meanings
    \item \textbf{High polysemy} (10+ senses): 870 lexemes (0.6\%)---highly frequent lexemes with domain-specific extensions
\end{itemize}

Highly polysemous lexemes include common terms like ``run'', ``set'', ``make'', ``take'', and ``get'', which exhibit broad semantic flexibility across contexts and domains.

\begin{figure}[t]
\centering
\includegraphics[width=0.8\columnwidth]{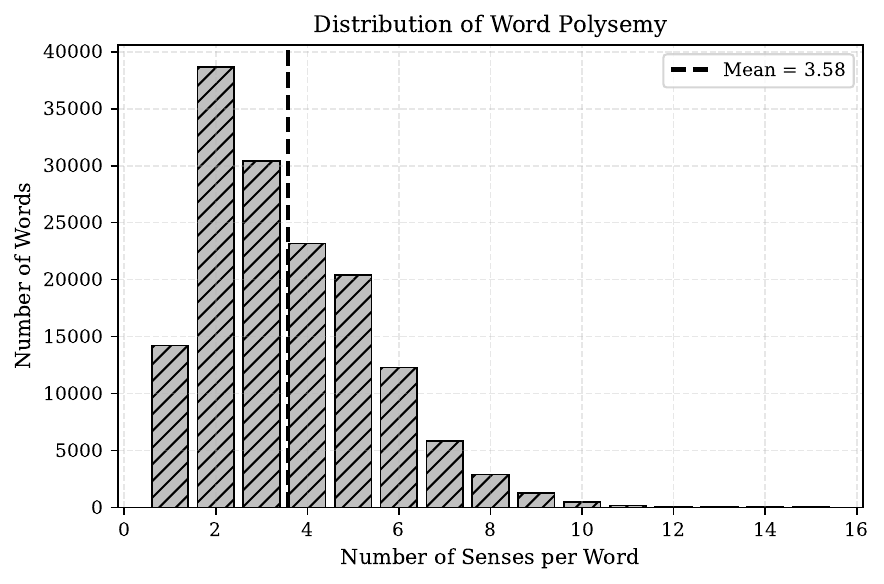}
\caption{Distribution of number of senses per lexeme in OpenGloss. Most lexemes (61.5\%) have 2-4 senses, with a long tail of highly polysemous lexemes reaching up to 24 senses. The median is 3 senses per lexeme.}
\label{fig:sense_distribution}
\end{figure}

Figure~\ref{fig:sense_distribution} visualizes the distribution of senses per lexeme, showing the characteristic long-tail pattern where most lexemes have 2-4 senses but a substantial minority exhibits extensive polysemy. This distribution balances sense granularity with practical utility: too few senses miss important distinctions, while excessive sense splitting creates spurious ambiguity.

\subsection{Semantic Graph Structure}
\label{sec:semantic-graph}

The semantic graph connecting OpenGloss senses and lexemes forms a rich network enabling traversal, reasoning, and semantic similarity computation. This subsection analyzes the graph's structure, connectivity patterns, and relationship distribution.

\paragraph{Edge Type Distribution}
The semantic network's 9.14 million edges span sense-level paradigmatic relations and POS-level syntagmatic patterns, capturing multiple dimensions of lexical semantics. At the sense level, synonymy dominates with 1.60 million edges (30.8\%), followed by hyponymy (27.3\% - 1.42 million edges), antonymy (21.6\% - 1.12 million edges), and hypernymy (20.3\% - 1.06 million edges). This distribution reflects the generative process: agents generate multiple synonyms per sense, create taxonomic hierarchies through hypernym/hyponym chains, and identify antonyms where semantically appropriate.

At the POS level, collocations account for 77.7\% (3.06 million edges), while inflections represent 22.2\% (875,673 edges). The dominance of collocations reflects their ubiquity in natural language: most content words participate in multiple conventional multi-word expressions. Table~\ref{tab:relationship_types} presents the distribution across the primary relationship types with counts and percentages.

Coverage analysis reveals near-universal connectivity that distinguishes OpenGloss from sparser semantic networks. Relationship coverage is comprehensive: synonym relations appear in 99.7\% of senses (mean 3.0 per sense), hypernyms in 99.9\% (mean 2.0), hyponyms in 98.6\% (mean 2.6), antonyms in 94.0\% (mean 2.1 when present), and usage examples in 99.7\% (mean 2.0).

This comprehensive relationship coverage ensures nearly every concept is embedded in a rich semantic neighborhood supporting similarity computation, taxonomy traversal, and conceptual reasoning.

\begin{table}[t]
\centering
\caption{Semantic Relationship Distribution}
\label{tab:relationship_types}
\begin{tabular}{lrrrr}
\toprule
\multirow{2}{*}{\textbf{Relationship Type}} & \multicolumn{2}{c}{\textbf{Sense-Level}} & \multicolumn{2}{c}{\textbf{POS-Level}} \\
\cmidrule(lr){2-3} \cmidrule(lr){4-5}
 & \textbf{Count} & \textbf{\%} & \textbf{Count} & \textbf{\%} \\
\midrule
Antonym & 1,123,954 & 21.6\% & --- & --- \\
Collocation & --- & --- & 3,063,419 & 77.8\% \\
Hypernym & 1,055,451 & 20.3\% & --- & --- \\
Hyponym & 1,419,971 & 27.3\% & --- & --- \\
Inflection & --- & --- & 875,673 & 22.2\% \\
Synonym & 1,599,958 & 30.8\% & --- & --- \\
\midrule
\textbf{Total} & \textbf{5,199,334} & \textbf{100.0\%} & \textbf{3,939,092} & \textbf{100.0\%} \\
\bottomrule
\end{tabular}
\end{table}

\paragraph{Graph Connectivity}
The semantic graph exhibits high and relatively uniform connectivity with minimal isolated nodes. Sense-level connectivity averages 17.0 edges (median 17, max 33, SD 2.6), while word-level connectivity averages 55.7 edges (median 49, max 330, SD not reported). Fewer than 0.1\% of senses lack semantic relationships. The low standard deviation in sense connectivity (2.6) indicates uniform coverage rather than the highly variable density of manually curated resources, facilitating graph traversal algorithms. Higher word-level variance reflects polysemy: more senses accumulate more edges through sense-level relationships, morphological variants, and collocations.

\subsection{Encyclopedic and Etymological Content}
\label{sec:encyclopedic-etymological}

Beyond lexicographic definitions and semantic relationships, OpenGloss provides two types of enrichment content that distinguish it from traditional computational lexicons: encyclopedic entries and etymological trails.

\paragraph{Encyclopedia Entries}
Encyclopedic content appears in 99.7\% of lexeme entries and 99.7\% of senses, providing 200-400 words of contextual explanation including conceptual significance, key characteristics, historical development, and domain-specific usage. Entries range from 180 to 420 words (mean 287, median 295), offering substantially more context than dictionary glosses while remaining more focused than full encyclopedia articles. This near-universal coverage distinguishes OpenGloss from resources offering only definitions (WordNet) or encyclopedia articles for prominent concepts only (Wikipedia via BabelNet).

\paragraph{Etymology Trails}
Etymology trails document historical development for 97.3\% of lexemes and 97.5\% of senses, tracing lexemes through historical languages, semantic evolution, cognates, morphological decomposition, and citations to etymological scholarship. The 97.5\% coverage (stopwords excluded) supports applications in morphological analysis, historical linguistics, and educational technology. While LLM-generated etymologies lack the authority of expert historical linguistics, they provide plausible developmental narratives useful for pedagogical purposes.

\paragraph{Definition Length Analysis}
Definition lengths span a typical range for lexicographic practice. Definition lengths reveal a median of 110 characters (mean 115 characters), exhibiting a characteristic right skew. This aligns with traditional dictionary practice: definitions are concise enough for quick reference but detailed enough for clear sense distinction. The distribution shows a characteristic right skew where most definitions are 60-140 characters, with a long tail of longer definitions for complex technical concepts requiring extended explanation.

\section{Comparative Analysis}
\label{sec:comparative_analysis}

Having established OpenGloss's scale and characteristics, we now compare it with major lexical-semantic resources: WordNet~\cite{miller1995wordnet,fellbaum1998wordnet}, BabelNet 5.3~\cite{navigli2012babelnet}, and ConceptNet 5.7~\cite{speer2012conceptnet,speer2017conceptnet}. We primarily compare against WordNet 3.0 (2006), the version distributed via NLTK and used in most NLP applications, while acknowledging that Princeton WordNet 3.1 (2011, online-only) and Open English WordNet 2024 (152K words, 121K synsets) continue development. Table~\ref{tab:comparison} presents comprehensive statistics across seven dimensions.

\begin{table}[t]
\small
\setlength{\tabcolsep}{4pt}
\begin{center}
\begin{tabular}{@{}lccccccc@{}}
\toprule
\textbf{Resource} & \textbf{Ver.} & \textbf{Year} & \textbf{Synsets} & \textbf{Edges} & \textbf{Enc.} & \textbf{Ety.} & \textbf{Lang.} \\
\midrule
WordNet & 3.0 & 2006 & 118K & -- & \xmark & \xmark & 1 \\
BabelNet & 5.3 & 2023 & 23M & -- & \cmark & \xmark & 600 \\
ConceptNet & 5.7 & 2021 & 8M & 21M & \xmark & \xmark & 83 \\
\textbf{OpenGloss} & \textbf{1.0} & \textbf{2025} & \textbf{537K} & \textbf{9.1M} & \cmark & \cmark & \textbf{1} \\
\bottomrule
\end{tabular}
\end{center}

\vspace{0.3cm}

{\sloppy
\begin{center}
\begin{tabular}{@{}lp{13.5cm}@{}}
\toprule
\textbf{Resource} & \textbf{Method \& Key Characteristics} \\
\midrule
\textbf{WordNet} & Manual curation by lexicographers; foundational resource for computational linguistics; \\
& Princeton's last release v3.0 (2006); v3.1 (2011) online-only; Open English WordNet 2024 (152K words, 121K synsets) continues development \\
\addlinespace[0.5ex]
\textbf{BabelNet} & Integration-based approach combining WordNet, Wikipedia, and machine translation; \\
& massive multilingual coverage but limited lexical precision per language \\
\addlinespace[0.5ex]
\textbf{ConceptNet} & Commonsense knowledge graph from crowd-sourced data and structured resources; \\
& emphasis on conceptual relations rather than lexical precision \\
\addlinespace[0.5ex]
\textbf{OpenGloss} & \textbf{LLM-generated encyclopedic dictionary} using multi-agent structured generation; \\
\textbf{(This work)} & Unique combination: lexical precision (537K senses) + encyclopedic content (200--400 words) \\
& + etymology trails (97\% coverage) + rich semantic network (99.9\% hypernym coverage) \\
\bottomrule
\end{tabular}
\end{center}
}

\vspace{0.2cm}

\footnotesize
\textbf{Notes:} \cmark~= included; \xmark~= not included; K = thousands; M = millions. Comparison uses WordNet 3.0 (147K words), the version distributed via NLTK and used in most NLP applications. OpenGloss (150K words) achieves 4.59$\times$ WordNet's synset count while sharing only 38\% vocabulary overlap, demonstrating high complementarity. OpenGloss uniquely combines encyclopedic content and etymology with lexical definitions. ConceptNet's edges represent commonsense relations; OpenGloss edges capture lexical-semantic relations (synonymy, antonymy, hypernymy, hyponymy) plus collocations and inflections. Edge counts for WordNet and BabelNet not specified in their documentation.

\caption{Comparison of OpenGloss with Major Lexical-Semantic Resources}
\label{tab:comparison}
\end{table}

\subsection{Scale and Coverage}
\label{sec:scale_coverage}

OpenGloss offers 4.59 times the synset coverage of WordNet 3.0 (536,829 vs. 117,659 synsets) while maintaining comparable lexicographic structure and moderate quality as assessed by automated QA (Section~\ref{sec:methodology:qa}). Critically, OpenGloss achieves comparable vocabulary breadth to WordNet (150,101 vs. 147,306 words, or 1.02$\times$), delivering both comparable word coverage and deeper semantic density than the manually curated gold standard while avoiding the dependency on pre-existing sources that constrains integration-based approaches.

\textbf{Coverage distinction:} The 4.59$\times$ comparison refers to synsets (sense definitions), not unique words. At the word level, OpenGloss now contains 150,101 unique lexemes compared to WordNet's 147,306---achieving 1.02$\times$ the unique word count (102\% of WordNet's vocabulary) with 3.58 senses per word on average. This milestone reflects OpenGloss's evolution beyond its initial focus on denser semantic coverage of common vocabulary to now achieving both breadth and depth. Measured against the wamerican dictionary (102,485 words), OpenGloss achieves 71.4\% coverage (73,200 words) compared to WordNet's 37.1\% (37,999 words), indicating substantially stronger alignment with everyday English vocabulary.

\textbf{Complementary vocabularies:} While OpenGloss now exceeds WordNet in total word count, the two resources prove highly complementary rather than redundant. Only 56,637 words appear in both resources (37.7\% of OpenGloss, 38.4\% of WordNet), while each contributes substantial unique vocabulary: 93,444 OpenGloss-only words (62.3\%) and 90,669 WordNet-only words (61.6\%). This symmetric distribution demonstrates that the resources serve distinct though overlapping purposes.

The shared vocabulary---including foundational terms like \textit{algorithm}, \textit{bank}, \textit{computer}, \textit{democracy}, and \textit{education}---represents core English lexical items where both resources agree on inclusion while differing on sense granularity (discussed in Section~\ref{sec:sense_granularity}). OpenGloss's unique contributions reflect contemporary usage patterns (\textit{smartphone}, \textit{cryptocurrency}, \textit{machine learning}, \textit{social media}), educational priorities (\textit{mindfulness}, \textit{sustainability}, \textit{digital literacy}), and conversational multi-word expressions (\textit{a lot of}, \textit{take into account}). WordNet's unique vocabulary consists predominantly of specialized technical terminology (\textit{deuteranopia}, \textit{phosphorylation}, \textit{eigenvalue}), proper noun instances (\textit{Alessandro Manzoni}, \textit{Melvil Dewey}), and multi-word expressions using underscore formatting incompatible with OpenGloss's space-based representation (\textit{common\_factor}, \textit{rock\_band}). The near-symmetric contribution---with each resource providing approximately 60\% unique words---positions them as complementary tools: OpenGloss for contemporary educational content and broad everyday vocabulary, WordNet for specialized terminology and fine-grained semantic distinctions refined through decades of expert curation.

Where WordNet prioritizes manual curation for English depth, BabelNet pursues multilingual breadth through integration. With 23 million synsets spanning 600 languages, BabelNet achieves massive scale by harvesting Wikipedia and Wiktionary content but averages approximately 2 synonyms per language per synset. In contrast, OpenGloss maintains richer internal structure through systematic generation, with near-universal relationship and example coverage (Section~\ref{sec:dataset_statistics}). For English-specific applications requiring semantic depth, this approach delivers denser lexical detail and encyclopedic context, albeit with noisier automatically generated relationships than manually curated resources.

A fundamentally different design philosophy distinguishes ConceptNet from lexicographic resources. Its 21 million edges connecting 8 million nodes (1.5 million English) capture commonsense knowledge through relationships like ``UsedFor,'' ``Causes,'' and ``LocatedAt''---patterns of everyday reasoning rather than lexicographic precision. OpenGloss's systematic lexical-semantic relationships (synonymy, antonymy, hypernymy, hyponymy) serve complementary purposes: ConceptNet excels at commonsense inference, while OpenGloss targets lexical semantics and educational content.

Beyond structural differences, a critical temporal dimension separates OpenGloss from existing resources. Princeton WordNet's last downloadable release was in 2006 (version 3.0, still distributed via NLTK), with version 3.1 (2011) available only online. While Open English WordNet continues annual development (2024 edition: 152K words), manual curation inherently limits update frequency. ConceptNet 5.7 dates to 2021, and BabelNet 5.3 (December 2023) reflects update latencies inherent in integration-based approaches. OpenGloss's 2025 generation demonstrates that LLM-based synthesis can produce up-to-date resources without waiting for manual curation or source material updates.

\subsection{Sense Granularity: A Detailed Comparison}
\label{sec:sense_granularity}

To understand how OpenGloss's sense distinctions compare with WordNet's decades of expert lexicographic refinement, we examined representative words spanning the polysemy spectrum---from specialized technical terms to highly polysemous everyday vocabulary. Table~\ref{tab:sense_comparison} presents direct comparisons revealing systematic differences in granularity philosophy.

\begin{table}[t]
\centering
\small
\begin{tabular}{@{}lccccl@{}}
\toprule
\textbf{Lexeme} & \multicolumn{2}{c}{\textbf{OpenGloss}} & \multicolumn{2}{c}{\textbf{WordNet}} & \textbf{Pattern} \\
\cmidrule(lr){2-3} \cmidrule(lr){4-5}
& \textbf{POS} & \textbf{Senses} & \textbf{POS} & \textbf{Synsets} & \\
\midrule
\textit{algorithm} & 1N & 2 & 1N & 1 & Technical: OG finer \\
\addlinespace[0.1cm]
\textit{bank} & 1N, 1V & 7 & 2N, 1V & 18 & Moderate: WN 2.6$\times$ \\
\addlinespace[0.1cm]
\textit{love} & 1N, 1V & 7 & 2N, 1V & 10 & Moderate: comparable \\
\addlinespace[0.1cm]
\textit{run} & 1N, 1V & 6 & 2N, 1V & 57 & High: WN 9.5$\times$ \\
\addlinespace[0.1cm]
\textit{set} & 1N, 1V, 1A & 10 & 2N, 1V, 1A & 45 & High: WN 4.5$\times$ \\
\bottomrule
\multicolumn{6}{@{}l@{}}{\footnotesize N=noun, V=verb, A=adjective; numbers indicate distinct POS categories} \\
\multicolumn{6}{@{}l@{}}{\footnotesize OG=OpenGloss, WN=WordNet; pattern describes granularity relationship}
\end{tabular}
\caption{Sense granularity comparison between OpenGloss and WordNet across the polysemy spectrum. Words selected to represent technical terms (algorithm), moderate polysemy (bank, love), and high polysemy (run, set). OpenGloss's 1-4 sense constraint creates coarser but more pedagogically accessible sense inventories, while WordNet's fine granularity reflects decades of expert lexicographic refinement for research applications.}
\label{tab:sense_comparison}
\end{table}

Three distinct patterns emerge from this comparison, reflecting fundamentally different lexicographic approaches. Technical and academic terms like \textit{algorithm} receive finer-grained distinctions in OpenGloss (2 senses differentiating procedural methods from rule-based systems) compared to WordNet's single general definition. This reflects OpenGloss's educational focus: students benefit from understanding that algorithms can be conceptualized both as step-by-step procedures and as formal rule systems, a distinction that aids comprehension across computer science, mathematics, and logic curricula.

At moderate polysemy levels, both resources converge on comparable coverage while diverging in organizational principles. Among moderately polysemous words like \textit{bank} and \textit{love} (7-10 total senses), WordNet's 18 synsets for \textit{bank} include highly specific uses (``bank as an arrangement of similar objects in a row,'' ``bank as acting as the banker in gambling''), while OpenGloss consolidates related meanings into pedagogically coherent sense clusters. The sparse distribution reflects OpenGloss's 1-4 sense constraint per part-of-speech---a deliberate design choice prioritizing learner comprehension over lexicographic completeness.

The granularity gap widens dramatically for the most semantically flexible vocabulary. For \textbf{highly polysemous words} like \textit{run} and \textit{set}, WordNet's advantage becomes pronounced: 57 synsets versus 6 for \textit{run}, 45 versus 10 for \textit{set}. These words exemplify English's most semantically flexible vocabulary, where decades of expert curation have identified subtle but linguistically motivated distinctions. WordNet distinguishes \textit{run} meanings including ``stretch out over distance'' (The road runs through the valley), ``extend in time'' (The play ran for years), ``be operating'' (The engine runs smoothly), ``tend to'' (Hemophilia runs in the family), and dozens more. OpenGloss's coarser granularity captures core meanings but necessarily omits specialized extensions.

Rather than representing resource quality deficiencies, these granularity differences reflect contrasting lexicographic philosophies. WordNet, developed for computational linguistics research, prioritizes sense distinctions that enable precise word sense disambiguation and semantic analysis. Its fine granularity supports applications requiring subtle meaning differentiation. OpenGloss, designed for educational technology and general NLP applications, balances comprehensibility with coverage: 3-4 senses per word provide sufficient semantic structure for most applications while remaining cognitively manageable for learners and computationally tractable for downstream systems.

This sparser sense inventory yields tangible computational benefits for graph-based applications. WordNet's density---57 synsets for \textit{run}, 45 for \textit{set}---creates semantic graphs where time-space complexity can become prohibitive for certain algorithms. Graph traversal, shortest path computation, and community detection algorithms often exhibit polynomial or exponential complexity in graph size; reducing node and edge counts by factors of 5-10$\times$ yields dramatic performance improvements. For real-time applications requiring sub-second response times (interactive educational systems, conversational agents, rapid semantic similarity computation), OpenGloss's coarser granularity may be preferred over WordNet's exhaustive coverage. This represents a pragmatic engineering trade-off: accepting reduced semantic precision in exchange for computational tractability and deployment feasibility in resource-constrained environments.

What OpenGloss sacrifices in sense granularity, it recovers through content richness absent from WordNet. Every entry includes 200-400 word encyclopedic context explaining conceptual significance, applications, and real-world relevance. For \textit{algorithm}, the encyclopedia entry contextualizes algorithms' centrality to computer science, mathematics, and automated decision-making---pedagogical framing that transforms bare definitions into learning resources. For \textit{run}, encyclopedic content explains polysemy's linguistic basis and cognitive foundations. Such encyclopedic depth, combined with etymology trails (97.3\% coverage), positions OpenGloss as a complementary resource: where WordNet excels in lexicographic precision for research applications, OpenGloss provides educational depth and accessibility for learners and general NLP systems.

\subsection{Key Differences}
\label{sec:relationship_comparison}

OpenGloss's graph exhibits uniform density contrasting with the variable coverage of manually curated resources (Section~\ref{sec:dataset_statistics}). BabelNet achieves scale through integration but introduces noise from automatic mapping, while ConceptNet's commonsense relationships (``UsedFor,'' ``Causes'') serve different purposes than OpenGloss's lexical-semantic structure.

\subsection{Novel Contributions}
\label{sec:novel_contributions}

OpenGloss uniquely combines three content dimensions at scale: lexicographic definitions, encyclopedic content, and etymology trails. WordNet omits encyclopedic and etymological content; BabelNet and ConceptNet provide only partial coverage of these dimensions. LLM-based generation ensures currency and enables rapid updates compared to manual curation latencies.

\subsection{Lexicographic Philosophy: Pragmatism vs. Purity}
\label{sec:lexicographic_philosophy}

Automated quality assurance evaluation (Section~\ref{sec:methodology:qa}) reveals that OpenGloss, like WordNet, adopts a pragmatic rather than purist approach to lexicographic coverage. Both resources include entries that strict traditional dictionaries would exclude---deliberate design choices prioritizing usability over convention.

\textbf{Inflected Forms:} Traditional print dictionaries use base forms as headwords (e.g., ``run'' not ``running,'' ``good'' not ``better'') to minimize redundancy and conserve pages. However, WordNet includes inflected forms as searchable entries: ``running'' has 52 synsets (noun: ``the act of running,'' ``the state of being in operation''; verb: ``move fast by using one's feet,'' etc.), ``better'' has 50 synsets, ``dogs'' has 8 synsets. This serves computational applications where users query the forms they encounter in text, not base forms they must derive through morphological analysis. A non-expert searching for ``running'' may not know to look under ``run''; a learner encountering ``better'' may not recognize it as the comparative of ``good.''

OpenGloss follows this practice: 32.9\% of our quality assurance sample includes inflected forms. Our QA agent initially flagged these as structural errors under strict dictionary conventions, but investigation of WordNet 3.0 confirmed this matches WordNet's established approach. The trade-off is deliberate: accept some redundancy and morphological impurity in exchange for accessibility and practical usability.

\textbf{Proper Nouns:} Traditional dictionaries exclude most proper nouns (cities, people, organizations), treating them as encyclopedic rather than lexical entries. The Oxford English Dictionary includes proper nouns only when they have ``passed into common use'' (e.g., ``Xerox'' as a verb). WordNet includes proper nouns as ``instance'' synsets: ``London'' (instance of city), ``Einstein'' (instance of physicist), ``Aachen'' (instance of city in Germany). This supports applications requiring named entity understanding, geographical knowledge, and biographical context.

OpenGloss includes 5.9\% proper noun entries in our quality assurance sample, providing educational value for vocabulary learners (students learning English benefit from knowing ``London'' is a city, ``Einstein'' a physicist) and geographic applications. The QA agent flagged these as violations of lexicographic conventions, but they represent intentional design alignment with WordNet's computational lexicography approach.

\textbf{Semantic Relationships:} Our quality assurance identified semantic relationship issues in 53.0\% of entries (hyponyms), 33.7\% (antonyms), 30.0\% (hypernyms), and 29.4\% (synonyms). These rates appear concerning until contextualized. Manual review suggests the QA agent applies stricter semantic hierarchy standards than WordNet itself: it flags near-synonyms as incorrect hyponyms, demands narrow taxonomic precision (``dog'' must be hyponym of ``canine,'' not ``mammal''), and rejects pedagogically useful but loose associations as errors.

While genuine semantic errors exist---listing nouns as hypernyms of verbs represents clear mistakes requiring correction---many flagged relationships represent acceptable variations. WordNet itself relaxes strict taxonomic logic for usability: semantic relationships serve both computational precision and human comprehension. The high flagging rate reflects a mix of (1) genuine errors requiring correction, and (2) conservative QA standards exceeding WordNet's own practices. Distinguishing these categories through human expert review represents important future work (Section~\ref{sec:discussion}).

This analysis validates OpenGloss's design philosophy: following WordNet's precedent of favoring practical utility over lexicographic orthodoxy. Both resources accept redundancy, morphological flexibility, and semantic imprecision in exchange for comprehensive, user-friendly coverage. For applications requiring strict lexicographic adherence to base-form headwords and taxonomic precision, traditional dictionaries or manually curated subsets remain necessary. For educational technology, vocabulary learning, and general NLP applications, the pragmatic approach has proven essential---a lesson OpenGloss inherits from WordNet's three decades of successful deployment.

\section{Discussion and Conclusion}
\label{sec:discussion}

This work demonstrates that LLM-based synthesis can generate comprehensive lexical resources exceeding the scale of manual curation while maintaining quality suitable for educational and general NLP applications. We reflect on what we have learned, characterize \opengloss's strengths and limitations, discuss applications and future directions, and address legal and ethical considerations.

\subsection{Key Insights and Achievements}

This work demonstrates that structured LLM generation can create comprehensive lexical resources at scales matching traditional manual curation. \opengloss provides 537{,}000 sense definitions across 150{,}000 lexemes---comparable to WordNet 3.1 in vocabulary breadth while delivering 4.59$\times$ more sense definitions, all generated in under 96 hours for under \$1,000. The comparative analysis (Section~\ref{sec:comparative_analysis}) establishes that synthetic generation enables combinations of content richness impractical for manual approaches: integrated definitions, encyclopedic context, etymological histories, usage examples, collocations, and semantic relationships. Automated quality assurance (Section~\ref{sec:methodology:qa}) confirms the resource's suitability for educational and general NLP applications, with core content quality supporting vocabulary learning and reading comprehension tasks while semantic relationships provide opportunities for continued refinement through community validation and iterative regeneration.

The multi-agent \texttt{pydantic-ai} pipeline with type-safe schemas enables reproducible generation with transparent validation, addressing a critical gap in computational lexicography: the ability to maintain currency. Traditional resources update on decade-long cycles because manual curation cannot keep pace with language evolution; WordNet's last major release was 2011. \opengloss demonstrates that synthetic generation can support continuous updates matching the pace of language change, with modular architecture allowing targeted regeneration of specific content types or vocabulary domains as foundation models improve. This reproducibility extends beyond updates: individual research groups can adapt the methodology for specialized domains, additional languages, or custom pedagogical requirements without institutional infrastructure.

The work establishes that \opengloss and WordNet serve complementary rather than competing roles within the lexical resource ecosystem. Only 56{,}637 words (38\%) appear in both resources, with each contributing substantial unique vocabulary reflecting different design priorities and temporal contexts. WordNet provides specialized technical terminology and exhaustive sense granularity refined through expert curation; \opengloss offers contemporary usage patterns, educational vocabulary, and rich pedagogical content (examples, encyclopedias, etymology, collocations). This complementarity demonstrates that synthetic and manually curated resources address different application requirements, expanding rather than replacing the toolkit available to researchers and educators.

\subsection{Resource Characteristics and Appropriate Use}

\opengloss exhibits a distinct quality profile shaped by its design philosophy and generation methodology. Sense granularity averages 3.58 per lexeme, constrained to 1--4 senses per part-of-speech---a deliberate design choice balancing learner comprehension with computational tractability. This produces coarser distinctions than WordNet's exhaustive coverage (WordNet's 57 synsets for ``run'' vs. 6 in \opengloss, 45 for ``set'' vs. 10), but enables more efficient graph traversal and real-time applications. The resource deliberately prioritizes educational and general-purpose terms over highly specialized technical terminology, reflecting its pedagogical focus.

As with any LLM-generated content, entries reflect patterns learned from training data rather than curated scholarly sources. Etymologies and encyclopedia content represent synthesized educational narratives; sense distinctions emerge from distributional patterns in language use. This makes \opengloss suitable for educational technology, prototyping NLP systems, and general vocabulary learning, but not authoritative reference material. For applications requiring precision, generated content should be validated against manually curated resources.

The resource necessarily reflects characteristics of training data, including potential underrepresentation of non-Western contexts, regional dialects, and diverse perspectives---areas for continued improvement across the entire field of language modeling. Users should be aware of these limitations when deploying \opengloss in educational or production systems.

\subsection{Applications}

\opengloss was originally designed for educational technology, and this pedagogical focus shaped both its content and structure. The combination of concise definitions with encyclopedic context (200--400 word entries explaining conceptual significance and real-world applications), usage examples demonstrating authentic word usage, etymological histories revealing word origins, and collocational patterns showing natural word combinations creates an integrated learning resource that goes beyond traditional dictionary lookup. Students can encounter new words through example sentences, explore conceptual depth through encyclopedia entries, understand word origins through etymology, and practice authentic usage through collocations. The same content enables automated content generation for K-12 curricula, adaptive vocabulary tutoring systems, and reading comprehension tools.

For computational linguistics research, word sense disambiguation has long relied on WordNet's semantic networks, but \opengloss offers a 4.59$\times$ larger sense inventory with near-universal relationship coverage (99.7\% of senses have synonyms, 99.9\% have hypernyms, 98.6\% have hyponyms, and 99.7\% include usage examples). The nearly 1.1 million example sentences (averaging 2 per sense) provide training data for context-sensitive disambiguation models. This expanded coverage provides richer semantic neighborhoods for disambiguation, particularly for polysemous terms that WordNet covers less comprehensively. Systematic evaluation on standard benchmarks like SemEval and Senseval corpora would quantify these potential advantages.

Knowledge-enhanced language models represent another natural application domain. Recent work on retrieval-augmented generation and knowledge-grounded systems has shown that explicit structured knowledge can improve factual consistency and reduce hallucination in neural language models. \opengloss provides exactly this kind of structured knowledge through its explicit taxonomies (hypernym/hyponym hierarchies organizing concepts from specific to general), synonym networks (grouping semantically similar terms), and encyclopedic content (grounding abstract concepts in explanatory context). While neural models already encode distributional semantics learned from text corpora, the explicit relationships in \opengloss offer complementary symbolic structure. Integration with systems like GraphRAG~\cite{edge2024graphrag} would merit investigation.

More broadly, \opengloss serves as a test bed for evaluating synthetic knowledge generation methods themselves. The resource enables direct comparison between different approaches to lexical knowledge acquisition: manual expert curation (WordNet's gold-standard precision), integration-based methods (BabelNet's massive multilingual coverage), crowdsourcing (ConceptNet's commonsense networks), and now systematic LLM-based generation with schema validation. Researchers investigating automated ontology learning, semantic similarity computation, or cross-lingual knowledge transfer can use \opengloss to quantify the trade-offs these different generation approaches entail.

These potential applications share a common requirement: systematic downstream evaluation remains critical for validating practical utility. Task-specific evaluation will reveal which applications benefit most from \opengloss's particular combination of scale, content richness, and structural consistency, and which require the precision of manually curated resources or domain-specific adaptation.

\subsection{Future Directions}

Immediate priorities include professional lexicographer evaluation---ideally 100--200 lexeme samples evaluated by multiple annotators with inter-annotator agreement metrics---to establish human expert consensus on content quality beyond automated QA. Benchmark evaluation on standard tasks would quantify performance against established baselines: word sense disambiguation (SemEval, Senseval datasets), semantic similarity (SimLex-999, WordSim-353), and lexical substitution (SemEval 2007 Task 10).

Expansion opportunities include multilingual contexts (Spanish, French, German, Chinese, Japanese building on the same generation methodology), domain-specific dictionaries (medical, legal, scientific, historical vocabularies), and alignment with established resources (WordNet, BabelNet, Wikidata, FrameNet, VerbNet) to leverage complementary strengths. Cross-lingual alignment and etymological networks spanning language families would provide unique value unavailable in monolingual resources.

The generation methodology supports continuous regeneration as foundation models improve, enabling incremental vocabulary additions tracking language evolution and version tracking documenting changes over time. We plan to open source the generation pipeline and codebase, enabling community replication, extension, and adaptation. Researchers and practitioners could customize schema designs, swap foundation models, or adapt prompting strategies for specific domains or languages.

We envision \opengloss not as a replacement for manually curated resources but as a complementary tool expanding what becomes computationally feasible in lexical knowledge engineering. This vision encompasses hybrid human-AI lexicography, where automated generation provides comprehensive coverage and rapid updates while human expertise establishes quality standards and validates critical entries. The reproducible methodology demonstrated here suggests pathways toward more dynamic, responsive lexical resources that can evolve alongside language itself.

\subsection{Legal and Ethical Considerations}

Foundation models train on massive text corpora that may include copyrighted dictionaries and encyclopedia articles, raising questions about whether generated content constitutes reproduction, derivative work, or independent creation. We implemented multiple safeguards: prompt engineering instructs models to generate original content without quoting existing dictionaries; schema constraints enforce length limits and structural requirements differing from typical dictionary formats; post-generation analysis found short n-gram matches with Wikipedia (n$\leq$10--15 words) but no passages exceeding 15 words reproduced verbatim, suggesting synthesis rather than memorization.

Under U.S. copyright law, facts themselves are not copyrightable---only creative selection and arrangement receive protection (\textit{Feist Publications, Inc.} v. \textit{Rural Telephone Service Co.}, 499 U.S. 340). Dictionary definitions may receive ``thin'' copyright if demonstrating originality, but basic sense distinctions (``a dog is a domesticated carnivorous mammal'') represent factual statements in the public domain. Recent scholarship analyzing fair use doctrine application to generative AI training~\cite{sag2024fairness,cooper2025files} and U.S. Copyright Office reports on AI and copyright~\cite{copyright_office2025} provide frameworks for assessing these issues, though legal consensus continues evolving.

We release \opengloss under CC-BY 4.0, requiring attribution but permitting commercial use, modification, and distribution. For academic research, educational applications, and general NLP development, these considerations are typically manageable under fair use doctrines. Commercial products requiring legal certainty should consult counsel and consider domain-specific validation against authoritative sources. The transparency of our generation methodology, automated quality assurance results, and open licensing aims to support informed decision-making about appropriate use contexts.

\subsection{Conclusion}

We opened this work with an ideal: lexical resources that bridge human comprehension and computational reasoning by integrating definitions, usage examples, encyclopedic context, etymology, and semantic relationships within a single schema. Existing resources face unavoidable trade-offs between quality, coverage, currency, and cost. This work demonstrates that systematic LLM-based generation with schema validation offers a new point in that design space.

\opengloss delivers on the three contributions outlined in the introduction. \textbf{First}, the resource provides 537{,}000 sense definitions across 150{,}000 lexemes with integrated content---encyclopedic context, etymological histories, usage examples, collocations, and 9.1 million semantic edges---addressing the gap between computational precision and pedagogical richness that existing resources leave unfilled. \textbf{Second}, the multi-agent generation pipeline with schema-validated outputs and automated quality assurance establishes a reproducible methodology accessible to individual research groups, produced in under 96 hours for under \$1{,}000 without institutional infrastructure. \textbf{Third}, empirical analysis reveals that \opengloss and WordNet are highly complementary rather than competitive, with only 38\% vocabulary overlap; each resource contributes distinct strengths, expanding rather than replacing the lexical resource ecosystem.

The resource addresses the trade-offs that have constrained lexicographic development. Where manual curation achieves exceptional quality at the cost of update latency (WordNet's last major release: 2011), \opengloss demonstrates that synthetic generation enables continuous updates matching language evolution. Where integration approaches achieve multilingual breadth but inherit schema inconsistencies, systematic generation produces uniform structure across all entries. Where crowdsourcing reduces cost but struggles with quality control, schema validation and automated QA provide transparent quality profiles. The result expands what becomes computationally feasible: comprehensive lexical resources that balance scale, consistency, currency, and cost in new ways.

We expect \opengloss to serve educators developing vocabulary learning tools, NLP researchers requiring large-scale semantic resources, and computational lexicographers investigating synthetic knowledge generation. The publicly available dataset and methodology under CC-BY 4.0 invite community participation in validating, extending, and adapting this work for diverse applications and languages. As foundation models continue improving, the reproducible generation methodology demonstrated here suggests pathways toward more dynamic lexical resources that can evolve alongside language itself---not replacing the precision of manual curation, but expanding the toolkit available to researchers and educators pursuing the ideal with which we began: every dictionary entry as an opportunity to bridge human understanding and machine reasoning.

{\sloppy
	\emergencystretch=2em
	\bibliographystyle{plain}
	\bibliography{bibtex/references}
}

\appendix
\section{Quality Assurance Details}
\label{appendix:qa}

This appendix reports detailed automated quality assurance (QA) results for the 1,000-entry validation sample described in Section~\ref{sec:methodology:qa}. The evaluation was designed to assess OpenGloss against our core design philosophy: emulating WordNet's pragmatic lexicographic approach that prioritizes computational utility over traditional conventions.

Table~\ref{tab:qa_results} presents a comprehensive quality profile organized into three key dimensions: overall quality assessment, analysis of flagged entries (distinguishing WordNet-aligned design choices from genuine improvement opportunities), and core content quality metrics.

\textbf{Key Findings:}

\textbf{1. Successful WordNet Replication (38.6\%):} The most significant finding confirms our design goals. Of the 688 flagged entries, 386 (38.6\%) were flagged specifically for including inflected forms (``running,'' ``dogs,'' ``better'') or proper nouns (``London,'' ``Einstein''). These are not quality failures but successful implementations of WordNet's established practices. As documented in Section~\ref{sec:lexicographic_philosophy}, WordNet deliberately includes these entry types to support computational applications where users query forms as encountered in text. Our conservative QA agent, applying traditional dictionary standards, flagged these as structural deviations---but they represent positive validation of our pragmatic design philosophy.

\textbf{2. Robust Core Content (62-79\%):} Core content quality proved strong across all dimensions: 79.0\% of entries showed no encyclopedia issues, 73.7\% no etymology issues, 65.3\% no example issues, and 62.4\% no definition issues. These rates reflect usable content suitable for educational and general NLP applications.

\textbf{3. Semantic Relationship Refinement Opportunities:} The primary area for improvement lies in semantic relationships, where 30-53\% of entries received flags for hypernym, hyponym, synonym, or antonym precision. As discussed in Section~\ref{sec:lexicographic_philosophy}, manual review suggests many of these flags reflect conservative QA standards that exceed WordNet's own practices---demanding stricter taxonomic precision than WordNet employs---rather than catastrophic errors. Distinguishing genuine errors from acceptable variations represents important future work.

\textbf{Methodological Note:} The 93.4\% overlap rate (most flagged entries have multiple flag types) reflects the QA agent's comprehensive evaluation across many dimensions. An entry flagged for including an inflected form might also receive a borderline hypernym flag---the categories are not mutually exclusive. This overlap pattern is expected when applying conservative standards across multiple quality dimensions simultaneously.

\begin{table}[h!]
\centering
\small
\begin{tabular}{@{}lrrl@{}}
\toprule
\textbf{Metric} & \textbf{Count} & \textbf{\%} & \textbf{Interpretation} \\
\midrule
\multicolumn{4}{@{}l@{}}{\textit{Overall Quality Profile}} \\
\addlinespace[0.05cm]
High Confidence & 141 & 14.1 & All criteria met without flags \\
Acceptable with Minor Issues & 171 & 17.1 & Core content usable, minor refinements \\
Flagged for Analysis & 688 & 68.8 & Specific, analyzable flags (see below) \\
\addlinespace[0.2cm]

\multicolumn{4}{@{}l@{}}{\textit{Analysis of Flagged Entries (688 total)}} \\
\addlinespace[0.05cm]
\multicolumn{4}{@{}l@{}}{\textbf{1. WordNet-Aligned Pragmatic Design (Positive Confirmation)}} \\
\addlinespace[0.03cm]
Inflected Forms & 329 & 32.9 & e.g., ``running,'' ``dogs,'' ``better'' \\
Proper Nouns & 59 & 5.9 & e.g., ``London,'' ``Einstein'' \\
\textbf{Either (union)} & \textbf{386} & \textbf{38.6} & \textbf{Successful WordNet replication} \\
\addlinespace[0.1cm]
\multicolumn{4}{@{}l@{}}{\footnotesize \textit{Note: These reflect deliberate design choices, not quality failures.}} \\
\addlinespace[0.2cm]

\multicolumn{4}{@{}l@{}}{\textbf{2. Opportunities for Iterative Refinement}} \\
\addlinespace[0.03cm]
Hyponym Relationship Flags & 530 & 53.0 & Taxonomic precision opportunities \\
Hypernym Relationship Flags & 300 & 30.0 & Hierarchical structure refinement \\
Antonym Relationship Flags & 337 & 33.7 & Oppositional relationship review \\
Synonym Relationship Flags & 294 & 29.4 & Meaning equivalence precision \\
Core Content Flags & 376 & 37.6 & Definitional or encyclopedic improvements \\
\addlinespace[0.1cm]
\multicolumn{4}{@{}l@{}}{\footnotesize \textit{Note: Categories overlap; 93.4\% of flagged entries have multiple flag types.}} \\
\multicolumn{4}{@{}l@{}}{\footnotesize \textit{Many semantic flags reflect conservative QA standards (see Section~\ref{sec:lexicographic_philosophy}).}} \\
\addlinespace[0.2cm]

\multicolumn{4}{@{}l@{}}{\textit{Core Content Quality (entries meeting standards)}} \\
\addlinespace[0.05cm]
No Definition Issues & 624 & 62.4 & Clear, precise, complete definitions \\
No Encyclopedia Issues & 790 & 79.0 & Factually accurate, pedagogically sound \\
No Etymology Issues & 737 & 73.7 & Plausible, educationally valuable \\
No Example Issues & 653 & 65.3 & Natural usage, appropriate context \\
\textbf{All Core Content Clean} & \textbf{271} & \textbf{27.1} & \textbf{No content quality flags} \\
\bottomrule
\end{tabular}
\caption{Quality assurance profile of 1,000 randomly sampled OpenGloss entries, evaluated against WordNet-aligned pragmatic standards. Key finding: 38.6\% of entries were flagged for including inflected forms or proper nouns---not quality failures, but successful implementation of WordNet's established practices. Core content quality is robust (62-79\% clean across dimensions), with semantic relationships representing the primary area for iterative refinement.}
\label{tab:qa_results}
\end{table}

\end{document}